\pdfoutput=1

\documentclass[11pt]{article}

\usepackage[final]{acl}

\usepackage{times}
\usepackage{latexsym}
\usepackage{booktabs}
\usepackage[T1]{fontenc}

\usepackage[utf8]{inputenc}
\usepackage{CJKutf8}

\usepackage{microtype}

\usepackage{inconsolata}

\usepackage{graphicx}
\usepackage[table]{xcolor}
\usepackage{colortbl}
\usepackage{amsmath, amssymb}
\usepackage{multirow}
\DeclareMathOperator*{\argmax}{arg\,max}

\title{Whisper-UT: A Unified Translation Framework for Speech and Text}

\author{
 \textbf{Cihan Xiao\textsuperscript{1}},
 \textbf{Matthew Wiesner\textsuperscript{1, 2}},
 \textbf{Debashish Chakraborty\textsuperscript{2}},
 \textbf{Reno Kriz\textsuperscript{2}},
\\
 \textbf{Keith Cunningham\textsuperscript{3}},
 \textbf{Kenton Murray\textsuperscript{1, 2}},
 \textbf{Kevin Duh\textsuperscript{1, 2}},
 \textbf{Luis Tavarez-Arce \textsuperscript{2}},
\\
 \textbf{Paul McNamee\textsuperscript{2}},
 \textbf{Sanjeev Khudanpur\textsuperscript{1, 2}}
\\
\\
 \textsuperscript{1}Center for Language and Speech Processing, Johns Hopkins University\\
 \textsuperscript{2}Human Language Technology Center of Excellence, Johns Hopkins University\\
 \textsuperscript{3}Georgetown University
\\
 \small{
   \textbf{Correspondence:} \href{mailto:cxiao7@jhu.edu}{cxiao7@jhu.edu}
 }
}

\begin{document}
\maketitle
\begin{abstract}
Encoder-decoder models have achieved remarkable success in speech and text tasks, yet efficiently adapting these models to diverse uni/multi-modal scenarios remains an open challenge. In this paper, we propose Whisper-UT, a unified and efficient framework that leverages lightweight adapters to enable seamless adaptation across tasks, including a multi-modal machine translation (MMT) task that explicitly conditions translation on both speech and source language text inputs. By incorporating ASR hypotheses or ground-truth transcripts as prompts, this approach not only enables the system to process both modalities simultaneously but also enhances speech translation (ST) performance through a 2-stage decoding strategy. We demonstrate our methods using the Whisper model, though in principle they are general and could be applied to similar multitask models. We highlight the effectiveness of cross-modal and cross-task fine-tuning, which improves performance without requiring 3-way parallel data. Our results underscore the flexibility, efficiency, and general applicability of the proposed framework for multi-modal translation.
\end{abstract}

\section{Introduction}
The task of speech-to-text translation (ST) encompasses converting spoken content from one language to another, aiming to overcome language barriers to communication. Traditionally, the task involves an automatic speech recognition (ASR) module to transcribe spoken words, followed by a machine translation (MT) module to convert the transcribed text into the target language in a cascaded manner~\cite{neyst}. The recent development of end-to-end neural architectures and large pre-trained models have substantially propelled advancements in downstream speech tasks, via either self-supervised learning (SSL)~\cite{wav2vec2, hubert, wavlm} or fully supervised learning. Among the pre-trained acoustic models, Whisper~\cite{whisper}, a transformer-based encoder-decoder multi-task model trained with large-scale data in a supervised manner, has exhibited good performance on various ST corpora.

However, in real-world scenarios, input modalities and data conditions vary widely. In offline settings, for instance, translating conversational or dialectal speech—characterized by disfluencies, code-switching, and noisy acoustic environments—poses significant challenges to end-to-end models, often resulting in degraded performance. Conversely, scenarios like business meetings or translated media archives frequently provide both source-language speech and (manual or ASR-generated) transcripts. Yet existing systems fail to exploit this multi-modal synergy.

To address this, we systematically investigate how multi-task encoder-decoder models—using Whisper as a representative case study—can be efficiently adapted to these heterogeneous scenarios. First, we examine fine-tuning strategies for conventional ST (using 3-way parallel speech-transcript-translation data), speech-to-text tasks (ASR-only data), and MT, while also methods for multi-modal translation where both speech and transcripts are available. Our analysis reveals two key insights:
\begin{itemize}
    \item \textit{Cross-task training induces synergistic benefits}—fine-tuning on in-domain ASR data improves ST performance, while ST training conversely enhances ASR accuracy, suggesting mutual reinforcement between the ASR and ST tasks even without 3-way parallel data;
    \item \textit{Multi-modal inputs (speech + text) consistently enhance translation quality when fused}, even with imperfect ASR transcripts.
\end{itemize}

Building on these findings, we propose \textbf{Whisper} for \textbf{U}nified \textbf{T}ranslation\footnote{We open source our code at \url{https://github.com/BorrisonXiao/Whisper-UT}.}, or \textbf{Whisper-UT}, a framework that transforms Whisper’s decoder into a unified conditional generation model, capable of dynamically conditioning on speech, text, or both modalities. The framework repurposes Whisper’s encoder-decoder architecture as a versatile multi-modal interface through two innovations:
\begin{itemize}
    \item [1] \textit{A multi-task learning paradigm} with a stochastic task-selection mechanism to adapt the system across ASR, MT, ST, and multimodal translation tasks using a single set of LoRA parameters;
    \item [2] \textit{A two-stage decoding strategy}, where the decoder first generates an ASR transcript from speech, then reuses it as context for translation, perhaps emulating human thought processes, even when a transcript is not provided.
\end{itemize}
Crucially, Whisper-UT requires no architectural modifications—only fine-tuning—ensuring compatibility with any encoder-decoder model.

Experiments on CoVoST2’s~\cite{wang2020covost2massivelymultilingual} French-English (\texttt{fr-en}) and German-English (\texttt{de-en}) subsets demonstrate strong performance. Extended evaluations on conversational telephone speech (CTS) corpora—Fisher-CallHome Spanish~\cite{fisher}, and BBN Mandarin-English~\cite{bbn} further confirm the robustness of our approach across diverse domains. Notably, Whisper-UT outperforms the 1.3B-parameter NLLB model in multi-modal settings (speech + ground-truth text) and achieves superior speech-only translation via hypothesis prompting.

Our work highlights the untapped potential of multi-task models in adaptive translation systems. By unifying modality handling and enabling efficient task specialization, Whisper-UT bridges the gap between rigid single-modality systems and the dynamic needs of real-world applications.

\section{Related Work}
\subsection{Whisper}
Whisper is an end-to-end multi-task speech model that adopts a transformer-like encoder-decoder architecture. Its \textsc{large-v2} version is pre-trained on 680,000 hours of speech data with multiple supervision.
As with the original transformer model~\cite{vaswani2023attention}, the loss function Whisper used at its pre-training time is the cross-entropy objective for all tasks.

Whisper's decoder supports a prompting mechanism, originally designed for better capturing long-range dependencies of the transcripts/translations to resolve local audio ambiguities. Particularly, long utterances are segmented into chunks and the decoder generates its hypothesis for the current segment conditioning on the previous segment's transcripts. Inspired by the effectiveness of GPT-like decoder-only models in machine translation, we hypothesize that Whisper's decoder, which may be viewed as an audio-conditional language model, is also capable of performing audio-augmented text generation conditioning on \textit{both inputs}. Our work extends recent work showing that the Whisper can be adapted via fine-tuning to perform a number of novel tasks including, audio-visual speech recognition \cite{rouditchenko2024whisper}, target-speaker ASR \cite{guo2024sq, polok2024target, ma2024extending}, translation to non-English languages \cite{peng2023prompting}, by showing that Whisper can be extended to enable multi-modal translation, i.e., using either only text or both text and speech inputs simultaneously.

\subsection{Multi-modal/-task Speech Systems}
Recent developments in multi-modal and multi-task systems, e.g.,~\cite{tang-etal-2021-ctfn}, are exploring new ways to combine audio and text to improve various language-related tasks. mSLAM~\cite{bapna2022mslam}, a multilingual speech and language model, has emerged as a pioneering approach. It aims to construct a shared representation space for both speech and text through joint pre-training on both self-supervised and supervised tasks with various loss objectives, including translation language modeling (TLM) loss for ST.

SeamlessM4T~\cite{seamlessm4t} is another innovative model that further refines the integration of multi-modal inputs for speech and text translation tasks. As a single model designed for ASR, T2T translation, T2S translation, S2T translation and S2S translation, it consists of multiple building blocks to leverage uni-modal data, including a w2v-BERT~\cite{chung2021w2vbert} as the speech encoder, a 1.3B NLLB model~\cite{nllb} as the text encoder and decoder, a transformer-based text-to-unit encoder-decoder model for speech, with a vocoder for converting the unit-sequences to waveforms.
These systems, along with most existing methods, primarily seek to simply align the representations of the text and speech modalities, limiting the model to still accept only one input modality at a time during inference, which prevents exploitation of \textit{cross-modal cues}.

More recently, speech‑centric large language models such as QWen‑Audio~\cite{chu2024qwen2audiotechnicalreport} have shown that a unified decoder can be fine‑tuned for a broad spectrum of text‑conditioned speech tasks—including contextual ASR~\cite{ragasr}—but these approaches rely on massive pretrained text LLMs and demand extensive data and compute during fine‑tuning. This is a gap we aim to fill.

A number of related works \cite{ma2024cross,zhang2023rethinking,liu2024recent,le2024comsl} have also demonstrated that multi-task learning can greatly improve speech translation performance.
Here, we focus on model fine-tuning and demonstrate that training end-to-end models for either ASR or ST alone improves performance on the other task, enabling fine-tuning with data that was not original annotated for the target domain task.   

\begin{figure}[!htbp]
  \centering
  \includegraphics[width=\linewidth]{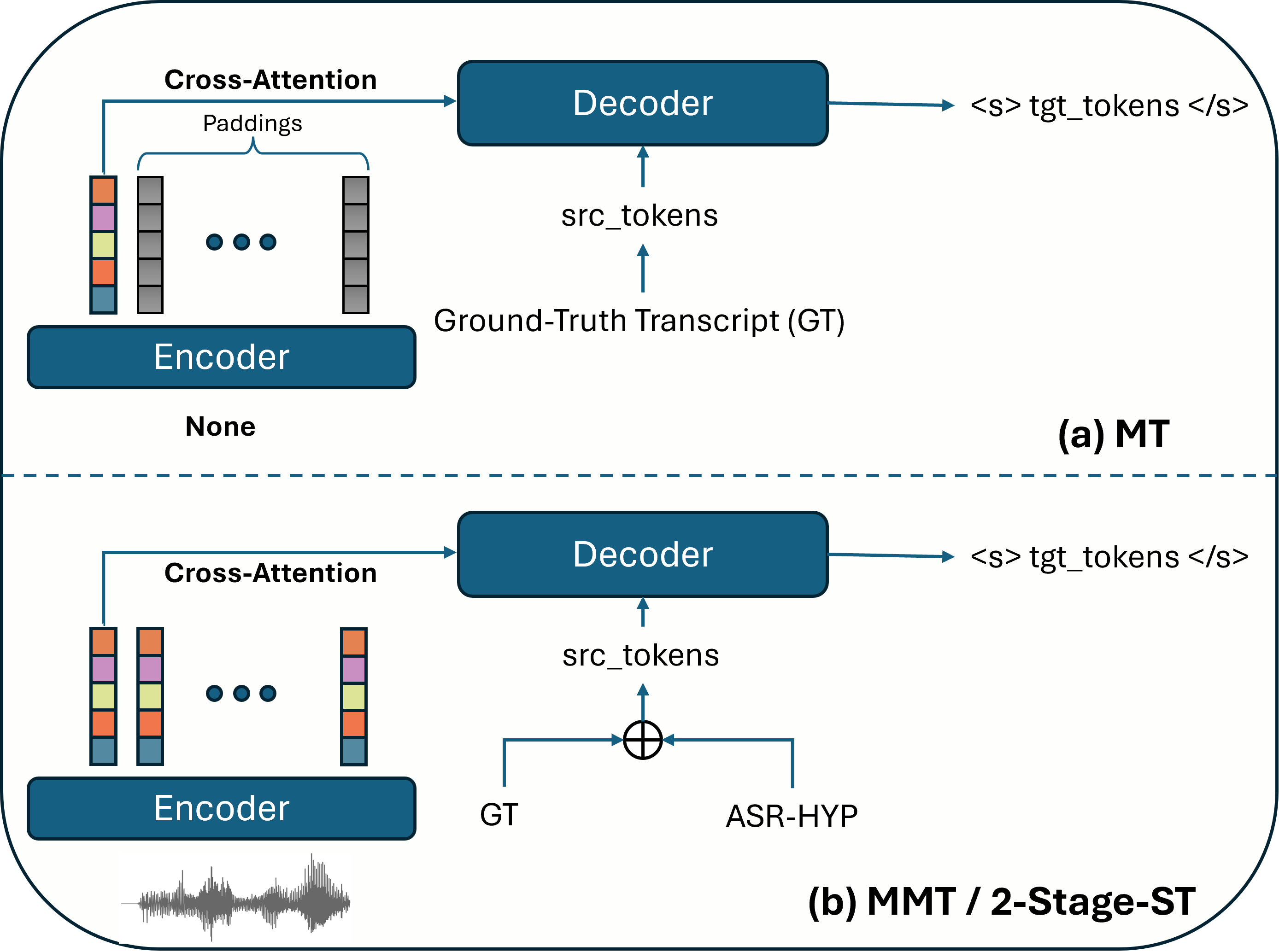}
  \caption{\textbf{Overview of our approach.} \textit{ASR-HYP} refers to the ASR hypothesis generated. When GT is used, the task is MMT, otherwise it is referred as 2-Stage-ST. The $\oplus$ symbol refers to the XOR operation. Note that special tokens are omitted to simplify illustration. }
  \label{fig:whisper-ut}
\end{figure}

\section{Methodology}
Traditional translation systems treat ST, MT, and ASR as distinct tasks, each requiring separate models or specialized architectures. In this work, we propose a \textbf{unified translation} framework that unifies these tasks under a single encoder-decoder paradigm, treating all forms of language conversion—including audio-to-text, text-to-text, and multi-modal translation—as conditional generation tasks. Our approach enables seamless adaptation to various input modalities and data conditions without requiring fundamental architectural changes.

At the core of our method is the insight that ASR can be reformulated as a source-language transcription task, ST as a direct speech-to-text translation task, and MT as a standard text-to-text translation task—all of which can be expressed as instances of sequence-to-sequence learning. Extending this idea, we introduce a \textbf{multi-modal translation} task, for which the model conditions on both speech and its corresponding transcript (either human-annotated or ASR-generated) to improve translation quality. This formulation generalizes the conventional ST and MT paradigms, leveraging available transcripts to enhance translation in scenarios where speech alone may be ambiguous or error-prone.

\subsection{Translation with Multi-modal Inputs}
We first provide a formal definition of the multi-modal translation 
(MMT) task, or more precisely, the task of speech-and-text-conditioned translation. Let $X = (x_1, x_2, \cdots, x_T)$ denote the speech signal of an utterance, $Y = (y_1, y_2, \cdots, y_M)$ denote the ground-truth transcript of the utterance, and $Z = (z_1, z_2, \cdots, z_N)$ denote its corresponding text translation. The goal of the task is then to find the conditional distribution $P(Z | X, Y)$. We hypothesize that often $H(Z | X, Y) < H(Z | Y)$ in practice, where $H$ denotes the information entropy.  In other words, the speech signal may contain additional information for a more accurate translation of the utterance, as it may be able to aid resolving ambiguities such as homographs, tonal variations, and omitted content—such as repetitions and filler words—that may be present in human-annotated transcripts.

In light of the remarkable performance observed with decoder-only language models in machine translation, we presume that encoder-decoder models' audio-conditioned decoder possesses the potential for undertaking the audio-conditioned text translation task. In particular, one may prompt the decoder with source language text, generated either by human annotators or any ASR system, in the translation process, as shown in Figure~\ref{fig:whisper-ut}(b). Consequently, the resulting model is trained to learn the distribution $P(Z | X, Y)$.

\subsection{Translation with Speech-only Inputs}
The problem of speech translation can be directly modeled as $P\left(Z | X\right)$ or modeled by marginalizing over an underlying latent variable, $Y^\prime$, representing valid transcripts of the audio $X$:
\begin{align}
    P(Z | X) &= \sum_{Y^\prime} P(Z, Y^\prime | X) \nonumber \\
    &= \sum_{Y^\prime} P(Z | Y^\prime, X) P(Y^\prime | X) \label{eq:st}
\end{align}
However, the summation over $Y^\prime$ is generally intractable. One common solution, also adopted by cascaded approaches to speech translation, is to approximate the summation with the single highest weight term in the summation, i.e.,
\begin{align*}
    \sum_{Y^\prime} P(Z | Y^\prime, X) P(Y^\prime | X) \\
    \approx \max_{Y^\prime} P(Z | Y^\prime, X) &P(Y^\prime | X),
\end{align*}
and furthermore to assume that the best transcript is the most likely one: 
\begin{align}
    \hat{Y} &= \argmax_{Y^\prime} P(Z | Y^\prime, X) P(Y^\prime | X) \nonumber \\
    &\approx \argmax_{Y^\prime} P(Y^\prime | X). \label{eq:argmax}
\end{align}

However, cascaded speech translation further assumes that the translation is conditionally independent of the audio given the transcript,
\begin{equation}
    P(Z|\hat{Y}, X) = P(Z | \hat{Y}),
\end{equation}
which is practical in that it enables modular training of components, i.e.,
\begin{equation}
    P(Z|X) = P(Z | \hat{Y})P(\hat{Y} | X),
\end{equation}
where $P(Z|Y)$ and $P(Y|X)$ can be trained separately, but it at the cost of a possible unneeded additional approximation.

End-to-end systems such as Whisper, however, model the problem without explicitly conditioning on the ASR transcripts, $Y^\prime$. Its single-decoder multi-task paradigm presumably captures a higher-level abstract semantics of the speech signals, such that the ST decoding process is implicitly entangled with the model's ASR ability.

We seek to combine the modeling advantages of the cascaded and end-to-end systems and generalize the multi-modal translation setting to re-formulate the system's speech-only translation process for approximating Equation~\ref{eq:st}. Specifically, we relax the conditional independence assumption of cascade approaches, by endowing end-to-end speech translation models with the capacity to also condition on either a ground-truth or hypothesized transcript defined by Equation~\ref{eq:argmax}, i.e.:
\begin{align}
    P(Z | X) &= P(Z | \hat{Y}, X) P(\hat{Y} | X)
\end{align}

In our implementation, we carry out a \textbf{two-stage decoding} process. In the first stage, the model is used to produce the ASR hypotheses, and subsequently, in the second stage, the model conditions on them to generate the translations.

An alternative perspective on this modeling is that it fully leverages the system’s source-language modeling capability. In end-to-end multi-task models, the decoder can be viewed as implicitly “partitioned” into two roles: source-language modeling and target-language generation. While these functions share parameters and benefit from joint optimization, they may still develop distinct competencies. By conditioning translation on both speech and textual transcripts, this approach explicitly harnesses a well-trained source-language model—potentially even from an external ASR system—allowing the decoder to generate more accurate translations. This perspective highlights how multi-modal conditioning can serve as a mechanism to refine and reinforce the system’s understanding of the source language, ultimately improving translation quality.

\subsection{Translation with Text-only Inputs}
Integrating MT functionality into a multi-modal encoder-decoder model presents unique challenges. In conventional encoder-decoder MT systems, the source language text is processed through the encoder, which generates contextual representations for the decoder to cross-attend to. However, oftentimes the pre-trained encoder is designed specifically for processing speech features, making direct text encoding potentially ineffective. Training the encoder to handle text inputs would require a significant amount of additional data and could lead to catastrophic forgetting, where the model loses its ability to process speech effectively.

Inspired by the success of decoder-only MT models such as GPT-like systems, we adopt an alternative strategy: instead of modifying the encoder to accommodate text, we encode the source text directly within the decoder, as illustrated in Figure~\ref{fig:whisper-ut}(a). Specifically, we prepend the source text as a prefix to the decoder input, leveraging the self-attention mechanism to implicitly model source-target dependencies. However, implementing this method within an encoder-decoder framework requires careful handling of the cross-attention mechanism. Since the decoder in our system is designed to attend to encoded speech representations, directly bypassing the encoder would disrupt the model's expected structure. To address this, we introduce a single learnable vector in the encoder, serving as an indicator that informs the decoder that text input is being processed. The remaining encoder output is padded with zeros, and we modify the cross-attention mask such that the decoder attends only to this learnable embedding. This design ensures that the model's architecture remains structurally intact while effectively repurposing the decoder for text-based translation.

\subsection{Whisper-UT: Unified Translation System}

To achieve a unified translation framework that encompasses multiple translation paradigms, we propose Whisper-UT, a system designed to handle ASR, ST, MT, and MMT within a single model. Our approach is built on multi-task learning, leveraging 3-way parallel data and text-only MT data to optimize multiple objectives in a stochastic fashion.

\subsubsection{3-way Parallel Data Objectives}
We formulate the learning process with six distinct training objectives, categorized based on the availability of parallel data.

For the 3-way dataset that provide speech, transcripts, and translations $\{X, Y, Z\}$, we define three primary objectives:

\noindent\textbf{ASR Objective.} Learning the mapping $X \to Y$, i.e., predicting the source language transcript from speech.

\noindent\textbf{E2E-ST Objective.} Directly predicting the target language text $Z$ from speech $X$.

\noindent\textbf{MMT Objective.} Predicting $Z$ while attending to both $X$ (speech) and $Y$ (source transcript).

\subsubsection{Text-Only Data Objectives}
Since 3-way parallel datasets are scarce in reality, we incorporate text-only MT data $\{Y, Z\}$ and define additional objectives:

\noindent\textbf{Source Language Modeling (SLM):} Predicting the next source token in $Y$, acting as an ASR surrogate for text-only samples.

\noindent\textbf{Target Language Modeling (TLM):} Predicting the next token in $Z$, improving the decoder's target language modeling ability.

\noindent\textbf{MT:} Translating $Y \to Z$.

For MMT and MT objectives, we allow gradients to propagate back through the source language tokens, implicitly enhancing the model's source language modeling ability.

\subsubsection{Dynamic Loss Weighting}
To balance the competing objectives, we employ a \textbf{stochastic task selection mechanism} with beta-distributed loss weighting inspired by \cite{modelmix}:
\begin{align}
    \alpha \sim \text{Beta}(\beta_1, \beta_2),
\end{align}
which determines the final multi-task loss:
\begin{align}
    \mathcal{L}_{\text{mtl}} &= (1 - \alpha) \mathcal{L}^{CE}_{\text{asr}} + \alpha \mathcal{L}^{CE}_{\text{st}}, \label{eq:loss}
\end{align}
where $\mathcal{L}^{CE}_{\text{asr}}$ is the ASR loss (or SLM loss for text-only samples), and $\mathcal{L}^{CE}_{\text{st}}$ is either the ST loss or the MMT loss, selected via stochastic task selection.

The stochastic weighting scheme is motivated by empirical findings that equal task weighting leads to gradient interference, degrading performance across tasks.

\subsubsection{Utterance-Level Task Selection}
Each batch is sampled from a mixture of the 3-way parallel data and text-only MT data. We define the loss computation as follows:
\begin{itemize}
    \item \textbf{ASR Loss:} Always computed for speech-based samples; replaced with SLM loss for text-only samples (zero-padded input except for a learnable vector).
    \item \textbf{ST vs. MMT Objective:}
With probability $q$, apply standard ST loss; for text-only data, this is equivalent to the TLM loss.
With probability $(1 - q)$, apply MMT loss, where the decoder cross-attends to both speech features and source text tokens; for text-only data, this becomes the conventional MT loss.
\end{itemize}

\subsubsection{Error Simulation in Multi-Modal Translation}
\label{sec:error}
For MMT, we introduce an ASR error simulation mechanism to enhance robustness. With probability $b$, we perturb a batch by replacing the source language tokens, sampled with probability $t$, with a similar alternative sampled randomly from the top-$k$ nearest neighbors in the embeddings space. To explicitly signal perturbed inputs, we prepend a special token to the modified sequence, allowing for the model to dynamically re-weight its reliance on the noisy text prefix and the corresponding audio input at inference time. This aims to simulate real-world noise in transcripts (e.g., ASR errors, omissions), encouraging the model to rely on both modalities for translation.

\subsection{Unified Training Framework}
\label{sec:ut}
In summary, our unified training framework integrates ASR, ST, MMT, and MT into a single multi-task learning process. To achieve this, we first concatenate both speech-text and text-only datasets, allowing for random sampling within each batch. For every batch, we compute the ASR loss, which corresponds to the source language modeling loss when dealing with text-only samples. The ASR and ST loss weights are dynamically balanced by sampling a weight $\alpha$ from a Beta distribution. Next, we stochastically determine whether the batch follows the ST/TLM objective or the MMT/MT objective. If the batch is selected for MMT training, ASR error simulation is applied with a certain probability to mimic transcription imperfections and enhance robustness. By combining these components, Whisper-UT serves as a unified model for ASR, ST, MT, and MMT, leveraging both textual and speech inputs efficiently.

\section{Experiments}

\subsection{Tasks and Datasets}
We test our approach on CoVoST2, a general-domain speech translation benchmark, using its French-English (180 hours) and German-English (119 hours) subsets for training. To assess performance on challenging conversational telephony speech (CTS), we conduct experiments on the Fisher-CallHome Spanish-to-English corpus (186 hours of spontaneous Spanish dialogues) and the BBN Mandarin-to-English corpus (110 hours of Mandarin-English telephony conversations). This setup tests our method’s adaptability across both general and domain-specific speech, with CTS posing unique challenges such as disfluencies, code-switching, and informal dialogue structures.

\subsection{Evaluation}
For both ASR and ST, we normalize the text by lower-casing all characters and removing all punctuations before computing the metrics. For the Fisher Spanish corpus, the BLEU score is computed with multiple references using the Moses~\cite{koehn-etal-2007-moses} toolkit as reported in other work~\cite{weiss2017sequencetosequence}. The evaluation script used is provided in the code.

\subsection{Training}
\label{sec:training}
To demonstrate our proposed approach, we adopt the \textsc{large-v2} version of Whisper with 1.6 billion parameters as the base model and fine-tune it for our unified translation modeling.
To enable joint training of speech-to-text and text-to-text translation within a single framework, we repurpose the 3-way parallel dataset by strategically replicating its text pairs. Specifically, we create a duplicate of the original dataset where the audio signals are removed, retaining only the source-target text pairs. This allows us to simulate text-only data without introducing external resources, ensuring parity in training scale across objectives.

\subsection{Experimental Results}

\begin{table}[!th]
\caption{
Direct Whisper fine-tuning results on the Fisher-Spanish and BBN-Mandarin datasets. The \textbf{Objective} column specifies under which training objective the model system is fine-tuned. \textit{None} refers to the original model. \underline{Underline} highlights the cross-task synergy.
}
\label{tab:synergy}
\centering
\addtolength{\tabcolsep}{-1pt}
\footnotesize
\begin{tabular}{c c| c | c c}
\toprule
& \textbf{Dataset} & \textbf{Objective} & \multicolumn{2}{c}{\textbf{Task}}\\
& & & ASR & E2E-ST\\
& & & (WER$\downarrow$) & (BLEU$\uparrow$)\\
\midrule
1 & \multirow{3}{*}{\textbf{Fisher}} & None & $26.7$ & $51.6$ \\
2 & & ASR & $19.1$ & \underline{$54.9$} \\
3 & & ST & \underline{$20.3$} & $61.2$ \\
\midrule
4 & \multirow{3}{*}{\textbf{BBN}} & None & $32.2$ & $13.0$ \\
5 & & ASR & ${18.9}$ & \underline{$16.2$} \\
6 & & ST & \underline{$23.1$} & ${16.8}$ \\
\bottomrule
\end{tabular}
\end{table}

\subsubsection{Overview}
Table~\ref{tab:synergy} presents results from directly fine-tuning Whisper, which reveals a cross-task synergy phenomenon: optimizing for one task (e.g., ASR) not only preserves but often enhances performance on another (e.g., ST), as indicated by underlined improvements across both datasets.
Table~\ref{tab:results} reports Whisper-UT results on three corpora: CoVoST2 (French $\rightarrow$ English, German $\rightarrow$ English), Fisher‑Spanish, and BBN‑Mandarin. Across all settings, our proposed Whisper‑UT variants demonstrate consistent improvements in transcription accuracy (WER↓) and translation quality (BLEU↑).

\begin{table*}[!th]
\caption{
Results on the test sets.
\textit{MMT} refers to the translation process that conditions on both the ground-truth transcript and the speech signals, while \textit{2-Stage-ST} refers to the MMT process with ASR hypothesis.
}
\label{tab:results}
\centering
\addtolength{\tabcolsep}{-1pt}
\footnotesize
\begin{tabular}{c c c c c c}
\toprule
& \textbf{Task} & \textbf{Dataset} & \textbf{Model} & \multicolumn{2}{c}{\textbf{Task}}\\
& & & & \textbf{Metrics} & \textbf{Results} \\
\midrule
1 & \multirow{12}{*}{ASR} & \multirow{3}{*}{\shortstack[*]{\textbf{CoVoST2}\\ \textit{fr-en} | \textit{de-en}}} & Baseline~\cite{wang2020covost2massivelymultilingual} & \multirow{3}{*}{WER$\downarrow$} & $18.3$ | $21.4$ \\
2 & & & Whisper-Large-V2 & & $13.4$ | $7.0$ \\
3 & & & Whisper-UT & & $\mathbf{8.3}$ | $\mathbf{5.8}$ \\
\cmidrule{3-6}
4 & & \multirow{6}{*}{\textbf{Fisher-Spanish}} & SeamlessM4T-Large & \multirow{6}{*}{WER$\downarrow$} & $76.3$ \\
5 & & & Whisper-Large-V2 & & $26.7$ \\
6 & & & Seq2seq~\cite{weiss17_interspeech} & & $23.2$ \\
7 & & & Multi-ASR~\cite{9003832} & & $22.9$ \\
8 & & & STAC-ST~\cite{zuluaga-gomez-etal-2023-end} & & $18.8$ \\
9 & & & Whisper-UT & & $\mathbf{16.3}$ \\
\cmidrule{3-6}
10 & & \multirow{3}{*}{\textbf{BBN-Mandarin}} & SeamlessM4T-Large & \multirow{3}{*}{WER$\downarrow$} & $52.6$ \\
11 & & & Whisper-Large-V2 & & $32.2$ \\
12 & & & Whisper-UT & & $\mathbf{17.4}$ \\
\midrule
13 & \multirow{8}{*}{MT} & \multirow{3}{*}{\shortstack[*]{\textbf{CoVoST2}\\ \textit{fr-en} | \textit{de-en}}} & Baseline~\cite{wang2020covost2massivelymultilingual} & \multirow{3}{*}{BLEU$\uparrow$} & $37.9$ | $28.2$ \\
14 & & & NLLB-1.3B & & $\mathbf{42.3}$ | $\mathbf{31.0}$ \\
15 & & & Whisper-UT & & $36.5$ | $26.9$ \\
\cmidrule{3-6}
16 & & \multirow{3}{*}{\shortstack[*]{\textbf{Fisher-Spanish}}} & NLLB-1.3B & \multirow{3}{*}{BLEU$\uparrow$} & $48.3$ \\
17 & & & Bi-NMT~\cite{9003832} & & $\mathbf{59.6}$ \\
18 & & & Whisper-UT & & $55.9$ \\
\cmidrule{3-6}
19 & & \multirow{2}{*}{\shortstack[*]{\textbf{BBN-Mandarin}}} & NLLB-1.3B & \multirow{2}{*}{BLEU$\uparrow$} & $8.7$ \\
20 & & & Whisper-UT & & $\mathbf{15.7}$ \\
\midrule
\multirow{2}{*}{21} & \multirow{4}{*}{MMT} & \multirow{2}{*}{\shortstack[*]{\textbf{CoVoST2}\\ \textit{fr-en} | \textit{de-en}}} & \multirow{2}{*}{Whisper-UT} & \multirow{2}{*}{BLEU$\uparrow$} & \multirow{2}{*}{$\mathbf{46.2}$ | $\mathbf{40.1}$} \\
\\
\cmidrule{3-6}
22 & & \textbf{Fisher-Spanish} & Whisper-UT & {BLEU$\uparrow$} & $\mathbf{70.4}$ \\
\cmidrule{3-6}
23 & & {\textbf{BBN-Mandarin}} & Whisper-UT &{BLEU$\uparrow$} & $\mathbf{26.0}$ \\
\midrule
24 & \multirow{18}{*}{ST} & \multirow{7}{*}{\shortstack[*]{\textbf{CoVoST2}\\ \textit{fr-en} | \textit{de-en}}} & Baseline~\cite{wang2020covost2massivelymultilingual} & \multirow{7}{*}{BLEU$\uparrow$} & $27.6$ | $21.0$ \\
25 & & & SeamlessM4T-Large & & $33.1$ | $35.8$ \\
26 & & & Whisper-Large-V2 & & $36.7$ | $36.8$ \\
27 & & & QWen2-Audio~\cite{chu2024qwen2audiotechnicalreport} & & $38.5$ | $35.2$ \\
28 & & & Whisper-UT & & ${40.8}$ | ${37.7}$ \\
29 & & & Whisper-UT-\textit{2-Stage} & & $\mathbf{41.4}$ | $\mathbf{38.1}$ \\
\cmidrule{3-6}
30 & & \multirow{7}{*}{\textbf{Fisher-Spanish}} & SeamlessM4T-Large & \multirow{7}{*}{BLEU$\uparrow$} & $14.7$ \\
31 & & & Multi-ST~\cite{9003832} & & $45.2$ \\
32 & & & Multi-task ST/ASR~\cite{weiss17_interspeech} & & $48.7$ \\
33 & & & Whisper-Large-V2 & & $51.6$ \\
34 & & & STAC-ST~\cite{zuluaga-gomez-etal-2023-end} & & $52.6$ \\
35 & & & Whisper-UT & & ${62.0}$ \\
36 & & & Whisper-UT-\textit{2-Stage} & & $\mathbf{62.1}$ \\
\cmidrule{3-6}
37 & & \multirow{4}{*}{\textbf{BBN-Mandarin}} & SeamlessM4T-Large & \multirow{4}{*}{BLEU$\uparrow$} & $7.0$ \\
38 & & & Whisper-Large-V2 & & $13.0$ \\
39 & & & Whisper-UT & & $19.8$ \\
40 & & & Whisper-UT-\textit{2-Stage} & & $\mathbf{21.6}$ \\
\bottomrule
\end{tabular}
\end{table*}

\subsubsection{Cross-task Synergy}
Table~\ref{tab:synergy} reveals that fine-tuning on one task does not only improve performance on the target task but also benefits other tasks as well. Notably, ASR fine-tuning enhances ST performance (51.6 to 54.9 on Fisher and 13.0 to 16.2 on BBN), and ST fine-tuning reciprocally benefits ASR (26.7 to 20.3 on Fisher and 32.2 to 23.1 on BBN).
This suggests that cross-task fine-tuning may mutually reinforce capabilities without architectural changes, inspiring Whisper-UT’s unified speech-text framework.

\subsubsection{ASR}
As shown in Table~\ref{tab:results}, on CoVoST2, Whisper‑UT reduces WER from 13.4/7.0 (Whisper) to 8.3/5.8. Similar gains appear on Fisher (from 18.8 to 16.3) and BBN (from 32.2 to 17.4). These improvements suggest that our stochastic task-interleaving mechanism effectively mitigates catastrophic forgetting, despite the addition of MT and MMT as new tasks. This stability preserves modality-specific expertise while introducing new tasks and enabling cross-task synergy.

\subsubsection{MT}
In text-only translation, Whisper-UT—trained without architectural modifications—narrowly trails the 1.3B-parameter NLLB model on general-domain CoVoST2 (36.5/26.9 vs. 42.3/31.0 BLEU) but surpasses it by +7.6 and +7.0 BLEU on domain-specific Fisher-Spanish (55.9 vs. 48.3) and BBN-Mandarin (15.7 vs. 8.7) benchmarks, despite using fewer parameters and no dedicated MT pretraining. This divergence highlights two key insights: (1) Whisper’s decoder inherently functions as a multilingual language model, capable of text-to-text translation with light-touch adaptation, and (2) its cross-lingual transfer capabilities, honed during speech-centric pretraining, generalize robustly to textual MT in low-resource, domain-specific scenarios. Critically, these results validate our hypothesis that minimal modifications—enabling joint training on speech and text—can unlock Whisper’s latent capacity for unified cross-modal translation, bridging the gap between speech and text without sacrificing architectural simplicity.

\subsubsection{MMT}
When translating with access to both speech and ground‐truth transcripts, Whisper‑UT achieves 46.2/40.1 BLEU on CoVoST2, 70.4 BLEU on Fisher‑Spanish, and 26.0 BLEU on BBN‑Mandarin—surpassing all MT baselines. This substantial improvement underscores the complementary nature of audio and text modalities: acoustic cues (e.g., prosody, emotion, pauses, repetitions) resolve ambiguities in noisy transcripts, while lexical context sharpens alignment of speech-derived semantics. By explicitly modeling these mutually compensatory signals, our unified architecture fuses audio and text modalities, yielding more robust translations when multi-modal information is available.

\subsubsection{ST}
In the ST setting, Whisper-UT achieves competitive performance with single-pass end-to-end decoding: 40.8/37.7 BLEU on CoVoST2 (\texttt{fr-en}/\texttt{de-en}), 62.0 BLEU on Fisher-Spanish, and 19.8 BLEU on BBN-Mandarin, surpassing QWen2-Audio, SeamlessM4T, and STAC-ST by margins of 2–8 BLEU points. Crucially, the 2-Stage inference variant yields systematic improvements over promptless decoding: +0.6/+0.4 BLEU on CoVoST2 (41.4/38.1 vs. 40.8/37.7), +0.1 BLEU on Fisher-Spanish (62.1 vs. 62.0), and +1.8 BLEU on BBN-Mandarin (21.6 vs. 19.8). These improvements are amplified in error-prone conditions, reflecting successful mitigation of ASR error propagation—a key challenge in cascaded systems. By prepending the special token during training (with simulated ASR noise) and inference (for 2-Stage decoding), the model learns to conditionally distrust imperfect transcripts while retaining their partial utility, rebalancing reliance on audio signals to correct latent errors. These consistent incremental gains validate the effectiveness of our two-stage modeling, demonstrating that even imperfect intermediate transcripts enhance translation fidelity through explicit cross-modal grounding when combined with learned distrust mechanisms.

\subsubsection{Summary}
The unified Whisper-UT framework achieves robust performance across three key tasks: monolingual ASR, text-only machine translation, and speech translation. Improvements are most pronounced in conversational Mandarin and Spanish settings. Moreover, the 2‑Stage decoding strategy provides a reliable way to enhance translation in fully end‐to‐end deployments. Overall, these results highlight Whisper-UT’s ability to unify cross-modal and cross-lingual speech-text tasks within a single architecture, offering a versatile solution for scenarios requiring joint speech-text modeling.

\section{Conclusion}
In this paper, we introduced Whisper-UT, a unified translation framework that integrates ASR, ST, MT, and MMT within a single multi-task learning paradigm. In addition to this unified framework, we propose an explicit modeling approach for speech translation that conditions on both speech signals and textual prompts, effectively leveraging ASR hypotheses or ground-truth transcripts. Our training strategy, incorporating stochastic task selection and modality-aware error simulation, ensures effective multi-task learning while mitigating catastrophic forgetting. Experimental results show that Whisper-UT achieves strong performance across various translation tasks, demonstrating the benefits of cross-task synergy. Future work will explore scaling to more languages and extending to broader multi-modal scenarios.

\vfill
\newpage

\section{Limitations and Ethical Considerations}
While our approach demonstrates strong improvements, several limitations remain. To ensure fair comparisons, we kept training steps consistent across models, meaning our best-performing system may not have reached its full potential with extended training. 

Due to resource constraints, we fine-tuned Whisper rather than training from scratch, which might limit the full integration of the objectives. Ideally, to demonstrate cross-task fine-tuning, we would start from a pretrained model that natively support each of our tasks, (MT, MMT, ST, ASR), but building state-of-the-art, or close to state-of-the-art systems requires building from existing models, such as Whisper, and adapting to Whisper to additionally perform these tasks, while a contribution in its own right, ultimately requires a two-stage fine-tuning approach that complicates analysis of the effectiveness of cross-task fine-tuning. Furthermore, while we believe our method to be general, i.e., it could be applied to similar models such as the OWSM model \cite{peng2024owsm}, we have only demonstrated our results using the Whisper model.

Training of machine learning models is a costly, energy-intensive process, so our method, which introduces a novel means of efficiently adapting existing large pre-trained models to new tasks, may mitigate the ethical concerns about the costs, financial, environmental, or other, associated with training ML models. Furthermore, the success of our approach, specifically cross-task fine-tuning, implies that speech translation systems can be more easily trained for new domains, including languages with limited training resources.

\vfill
\clearpage

\bibliography{custom}

\clearpage

\appendix

\section{Training Detail}
\subsection{Parameter Efficient Fine-tuning}
\label{sec:peft}
To efficiently adapt the model to these conversational scenarios without overfitting or incurring excessive computational cost, we leverage several parameter-efficient fine-tuning (PEFT) techniques.

In order to fit the base model into our hardware, we adopt a list of strategies:
\begin{itemize}
    \item \textbf{Low-Rank Adaptation (LoRA).} LoRA~\cite{hu2021lora} introduces a trainable adapter comprised of rank decomposition matrices on top of the fixed pre-trained model's weight matrices in specified layers so that the number of trainable parameters can be considerably reduced.
    \item \textbf{Gradient checkpointing.} Gradient checkpointing~\cite{chen2016training} stores intermediate activations in the forward pass, and re-computes the remaining activations during back-propagation.
    \item \textbf{Zero Redundancy Optimizer (ZeRO).} ZeRO~\cite{rajbhandari2020zero} is an algorithm that partitions data, optimizer states, gradients, and parameters for speeding up the training of large neural models with low communication costs.
\end{itemize}

\subsection{Hyperparameter Settings}
\label{sec:hyperparams}

Table~\ref{tab:hyperparams} presents the hyperparameter configurations used for training our Whisper-UT model.

\begin{table}[h]
    \centering
    \begin{tabular}{l c}
        \toprule
        \textbf{Hyperparameter} & \textbf{Value} \\
        \midrule
        LoRA Rank & 200 \\
        LoRA Alpha & 400 \\
        LoRA Dropout & 0.1 \\
        \midrule
        Max Training Steps & 10000 \\
        Batch Size & 64 \\
        Gradient Accumulation Steps & 1 \\
        Warmup Steps & 500 \\
        Learning Rate & $1\text{e}^{-5}$ \\
        Weight Decay & $5\text{e}^{-4}$ \\
        \midrule
        SpecAug Mask Feature Probability & 0.1 \\
        SpecAug Mask Time Probability & 0.05 \\
        \bottomrule
    \end{tabular}
    \caption{Hyperparameter configurations used for training.}
    \label{tab:hyperparams}
\end{table}

Experiments in this work are conducted with 8 V100-32GB GPUs. However, PEFT methods outlined in Section~\ref{sec:peft} render the use of 8 GPUs redundant, yet they are deployed to accelerate the training process.

\subsection{Data Augmentation}
We apply the conventional speed perturbation~\cite{ko15_interspeech} with parameters 0.9, 1.0, 1.1 to the speech prior to the training stage. Additionally, we adopt SpecAug~\cite{specaug} to randomly mask extracted speech features during training.

\section{CTS Data Detail}
\subsection{Pre-processing}
CTS corpora usually consist of short utterances segmented from a full recording, reflecting the alternating speech of participants during conversations. However, we found empirically that fine-tuning on such segments, presumably due to a mismatch in sample lengths compared to Whisper's pre-training data, leads to significant performance degradation. The resulting model tends to repetitively produce frequent filler words in the training corpus at inference time regardless of the input. Therefore, we re-segmented the utterances by merging them chronologically, with durations (in seconds) sampled from a Gaussian distribution, e.g. $\mathcal{N}(15, 5^2)$. As Whisper's feature extractor automatically pads the features up to 30 seconds, such re-segmentation also significantly reduced the training cost in terms of memory and time.

\begin{CJK}{UTF8}{gbsn}
\begin{table*}[!ht]
\centering
\caption{Code-switching example with system outputs.}
\begin{tabular}{ll}
\toprule
\textbf{REF-ASR:} & 电脑的 MASTER 应该是很 POPULAR 就对了很应该很 \\
\textbf{HYP-ASR:} & 电脑的 master 应该是很 popular 就对了很应该很 \\
\textbf{REF-MT:} & MASTER degree of computer science it should be very POPULAR it should be \\
\textbf{HYP-E2E-ST:} & The computer should be very popular, should be very \\
\textbf{HYP-2-Stage-ST:} & The computer's master should be very popular that's right very should be very \\
\textbf{HYP-MMT:} & The computer's MASTER should be very popular that's right very should be very \\
\bottomrule
\end{tabular}
\label{tab:code_switch}
\end{table*}

\subsection{BBN-Mandarin Data Specification}
The BBN Mandarin-English conversational telephony speech (CTS) corpus used in our experiments comprises two primary components:  
\begin{itemize}
    \item \textbf{HKUST Mandarin ASR Dataset} (90.1 hours): Mandarin conversational speech from telephony interactions, originally designed for ASR research~\cite{fung2005hkust}.
    \item \textbf{CallHome Mandarin ASR Dataset} (20.5 hours): Informal Mandarin dialogues curated for ASR study~\cite{canavan1996callhome}.
\end{itemize}
\noindent The BBN team~\cite{bbn} translated these into English to create parallel speech-to-text translation pairs. While our experiments utilized a pre-publication version provided directly by the BBN authors, minor discrepancies (e.g., data splits, preprocessing, or translation refinements) may exist compared to the final published version. Nevertheless, the corpus retains its core characteristics: conversational telephony domain focus, code-switching prevalence, and disfluency patterns.

\section{Qualitative Analysis of Code-Switching}
The code-switching example presented in Table~\ref{tab:code_switch} demonstrates two critical insights:

\begin{itemize}
    \item \textbf{ASR Preservation of Linguistic Salience:} The 2-Stage decoding system successfully retains the code-switched terms ``master'' and ``popular'' (WER $\approx$ 0\% for these tokens), while E2E-ST completely omits ``master''. This suggests that: 1) direct audio-to-translation mapping struggles with lexical disambiguation of homophones (``master'' vs. contextually expected ``computer''), and 2) explicit intermediate ASR provides discrete textual anchors that guide translation decisions.

    \item \textbf{Cross-Modal Faithfulness:} While the reference MT (REF-MT) omits the final ``很'' (translated as "very") from the source utterance ``很应该很'', our ASR transcript preserves all repetitions. This discrepancy highlights how audio-derived prosodic cues (e.g., emphatic stress on the final ``很'') enable 2Stage-ST and MMT to retain pragmatic emphasis (``...that's right very should be very'') where text-only MT truncates for conciseness. By aligning acoustic signals (stress patterns) with textual redundancy, our framework distinguishes intentional repetition—a discourse marker of conviction in Mandarin—from superficial noise, demonstrating superior faithfulness to both linguistic content and pragmatic intent compared to E2E ST pipelines.
\end{itemize}

The example validates our hypothesis that two-stage processing particularly benefits scenarios where: 1) ASR can reliably capture linguistically salient content (code-switches, proper nouns), and 2) Audio signals contain complementary paralinguistic information (prosodic boundaries, emphasis) that each modality alone cannot convey. This dual-modality advantage explains 2-Stage-ST's performance gain over E2E-ST on BBN-Mandarin despite identical model parameters.
\end{CJK}

\section{Ablation Study}

\begin{table*}[!th]
\caption{
Ablation studies on the CTS test sets. The \textbf{Objective} column specifies under which training objective the model system is fine-tuned. The \textit{UT} objective refers to the unified-translation objective described in section~\ref{sec:ut}. The \textbf{Task} column specifies the target inference task. \textit{E2E-ST} refers to the promptless E2E speech translation setting, \textit{MMT} refers to the translation process that conditions on both the ground-truth transcript and the speech signals, while \textit{2-Stage-ST} refers to the MMT process which conditions on the model's own ASR hypotheses.
}
\label{tab:ablation}
\centering
\addtolength{\tabcolsep}{-1pt}
\footnotesize
\begin{tabular}{c c | c | c | c c c c c}
\toprule
& \textbf{Dataset} & \textbf{Model} & \textbf{Objective} & \multicolumn{5}{c}{\textbf{Task} (num\_beams = 1)}\\
& & & & ASR & E2E-ST & MT & MMT & 2-Stage-ST\\
& & & & (WER$\downarrow$) & (BLEU$\uparrow$) & (BLEU$\uparrow$) & (BLEU$\uparrow$) & (BLEU$\uparrow$)\\
\midrule
1 & \multirow{11}{*}{\textbf{Fisher}} & \multirow{2}{*}{NLLB-1.3B} & None & -- & -- & $48.3$ & -- & --\\
2 & & & MT & -- & -- & $\mathbf{67.3}$ & -- & --\\
\cmidrule{3-9}
3 & & \multirow{8}{*}{Whisper} & None & $26.7$ & $51.6$ & -- & -- & --\\
4 & & & ASR & $19.1$ & $54.9$ & -- & -- & --\\
5 & & & ST & $20.3$ & $61.2$ & -- & -- & --\\
6 & & & ASR + ST & $16.3$ & $\mathbf{62.2}$ & -- & -- & --\\
7 & & & MT & $60.3$ & $51.0$ & $63.4$ & $61.1$ & $52.4$\\
8 & & & ASR + ST + MT + LM & $\mathbf{16.0}$ & $61.7$ & $55.2$ & -- & --\\
9 & & & MMT & $16.4$ & $57.4$ & $1.4$ & $67.5$ & $58.6$\\
10 & & & UT-OOD & $\mathbf{16.0}$ & ${61.5}$ & $44.2$ & $70.0$ & $61.6$\\
11 & & & UT-CTS & $16.3$ & $62.0$ & $55.9$ & $\mathbf{70.4}$ & $\mathbf{62.1}$\\
\midrule
12 & \multirow{11}{*}{\textbf{BBN}} & \multirow{2}{*}{NLLB-1.3B} & None & -- & -- & $8.7$ & -- & --\\
13 & & & MT & -- & -- & $\mathbf{22.7}$ & -- & --\\
\cmidrule{3-9}
14 & & \multirow{8}{*}{Whisper} & None & $32.2$ & $13.0$ & -- & -- & --\\
15 & & & ASR & $18.9$ & $16.2$ & -- & -- & --\\
16 & & & ST & $23.1$ & $16.8$ & -- & -- & --\\
17 & & & ASR + ST & $18.5$ & $20.2$ & -- & -- & --\\
18 & & & MT & $37.7$ & $12.7$ & $16.0$ & $20.4$ & $15.5$ \\
19 & & & ASR + ST + MT + LM & $17.7$ & $20.4$ & $14.8$ & -- & --\\
20 & & & MMT & $17.5$ & $19.5$ & $1.0$ & $25.2$ & $20.6$\\
21 & & & UT-OOD & $17.5$ & $\mathbf{20.6}$ & $11.1$ & $25.3$ & $21.5$ \\
22 & & & UT-CTS & $\mathbf{17.4}$ & $19.8$ & $15.7$ & $\mathbf{26.0}$ & $\mathbf{21.6}$\\
\bottomrule
\end{tabular}
\end{table*}

We conduct ablation experiments presented in Table~\ref{tab:ablation} on the two CTS datasets (Fisher-Spanish and BBN-Mandarin), as their domain-specific challenges—disfluencies, code-switching, and spontaneous dialogue—diverged significantly from Whisper’s pretraining data. This allows us to isolate our framework’s adaptability beyond pretraining biases and quantify its efficacy in resource-constrained, real-world scenarios.

\subsection{Text-only MT Training and Its Effects}  
Rows 7 and 17 show the results of the MT-only fine-tuning experiment, demonstrating that the model achieves strong text translation performance even with limited in-domain data—BLEU 63.4 on Fisher-Spanish and 16.0 on BBN-Mandarin. This outperforms the original NLLB-1.3B model, though it remains modestly behind its fine-tuned counterpart. This suggests that Whisper's decoder inherently possesses some text translation capabilities or at least has sufficiently strong source and target language modeling abilities such that minimal adaptation enables it to perform the MT task. Interestingly, this MT training also gives the system MMT ability, as suggested by the 61.1/20.4 (Fisher/BBN) BLEU score, despite MMT being a novel objective that the model was not explicitly trained on. In fact, on the BBN corpus, the MT-trained model exhibits MMT capabilities that surpass its original training objective, achieving a BLEU score of 20.4 (MMT) compared to 16.0 (MT). This finding reinforces our earlier observation of cross-task synergy.

\subsection{Effectiveness of Multi-task Learning}  
In rows 6 and 17, we conduct straightforward multi-task fine-tuning experiments by duplicating the speech dataset with both ASR and ST supervision, concatenating the datasets, and employing random sampling within each batch. These experiments confirm that multi-task training is beneficial, as it enhances BLEU score from 61.2 to 62.2 and WER is reduced from 20.3 to 16.3 on the Fisher-Spanish corpus. A similar trend is observed on the BBN set as well. This suggests that jointly optimizing multiple relevant objectives allows the model to better capture linguistic patterns and improve generalization across tasks.

\subsection{MMT-Multi-task Training and Its Implications}  
Rows 9 and 20 evaluate MMT-multi-task fine-tuned models, that is, the model is trained with $q = 0$ and $b = 0$. Notably, the MMT inference results outperform even the strong fine-tuned NLLB-1.3B baseline in MT performance, 70.4 vs. 67.4 on Fisher and 26.0 vs. 22.7 on BBN— demonstrating that MMT provides tangible benefits over traditional cascaded MT approaches.

However, a gap remains between different MMT settings. Specifically, when using the ASR hypothesis as input instead of the ground-truth transcript, i.e., the 2-Stage-ST decoding, performance drops from 67.5 to 58.6 on Fisher and from 25.2 to 20.6 on BBN. While this still exceeds the results from direct ST (52.4 vs. 51.0 on Fisher and 20.6 vs. 19.5 on BBN), the model tends to over-rely on the transcript in the absence of explicit modeling. Specifically, without the special tag to signal potential errors, the model treats the input transcript as fully reliable ground truth—an assumption that breaks down when using ASR outputs, which may contain recognition errors. These highlight both the effectiveness of explicit modeling and the limitations introduced by ASR errors.

\subsection{Unified Translation (UT) Training}
\subsubsection{Overview}
Finally, the UT-trained system (rows 11 and 22) achieves the best MMT and 2-Stage-ST results, with MMT reaching 70.4/26.0 BLEU and 62.1/21.6 BLEU, respectively, on the Fisher-Spanish and BBN-Mandarin corpora, proving the method's effectiveness. Applying the error simulation strategy in this training scheme improves the robustness of the two-stage approach, narrowing the performance gap between MMT and 2-Stage-ST decoding. Specifically, on Fisher, the gap decreases from 8.9 to 8.3 BLEU (row 9 vs. 11), and on BBN, from 4.6 to 4.4 BLEU (row 20 vs. 22), indicating more stable performance under ASR-transcript input.

\subsubsection{Analysis of Transcript-Conditioning}
On the Fisher test set, the 2‑Stage‑ST decoding strategy of the Whisper‑UT model actually falls slightly behind the simpler ASR+ST multi‑task E2E‑ST model. Direct multi‑task training of ASR and ST (row 6) achieves a BLEU of 62.2, whereas conditioning on ASR hypotheses under the unified‑translation objective (row 11, 2‑Stage‑ST) yields 62.1—a 0.1 BLEU drop. Through manual inspection, we found this gap is driven largely by inconsistent translation of filler words: the same Spanish filler (e.g., “eh,” “um”) in ASR transcripts is rendered inconsistently in output, magnifying ASR transcript “errors” during translation. Moreover, because Whisper’s ASR and ST performance on Fisher Spanish are both strong already (WER $\approx$ 16, BLEU $\approx$ 60), there is little mismatch for transcript conditioning to resolve, so the transcript signal offers marginal benefit.

In contrast, on the BBN corpus, the UT model demonstrates a clear advantage. The ASR+ST multi‑task E2E‑ST model (row 16) scores 20.2 BLEU, while the Whisper‑UT 2‑Stage‑ST decoder (row 22) jumps to 21.6 BLEU—a significant 1.4‑point gain. This larger benefit arises because BBN combines relatively low WER ($\approx$ 18) with much lower translation quality (BLEU $\approx$ 20), indicating that the model’s ST ability lags behind its ASR competence. In this scenario, explicitly leveraging ASR transcripts helps fill the performance gap, yielding more accurate translations under the unified objective.

\subsection{Impact of Out-of-Domain Text Data}
\subsubsection{Dataset Setup}
To evaluate the robustness of our unified framework to domain shifts in text data, we replace the in-domain machine translation (MT) pairs (derived from CTS audio transcripts, as described in Section~\ref{sec:training}) with out-of-domain (OOD) text pairs. Specifically:

\noindent\textbf{Spanish:} We use 197 hours of text pairs from three sources:
\begin{itemize}
    \item CoVoST 2~\cite{wang2020covost2massivelymultilingual} (diverse web-mined speech),
    \item mTEDx~\cite{salesky2021multilingualtedxcorpusspeech} (TED talk subtitles), and
    \item Europarl-ST~\cite{koehn-2005-europarl} (parliamentary proceedings).
\end{itemize}

\noindent\textbf{Mandarin:} We include 130 hours from:
\begin{itemize}
    \item CoVoST~\cite{wang2020covostdiversemultilingualspeechtotext} (multilingual web content),
    \item GALE~\cite{song2016gale} (broadcast news and interviews), and
    \item proprietary in-house datasets (mixed genres).
\end{itemize}

The OOD sets contrast sharply with CTS data in domain (e.g., formal talks vs. casual dialogues) and lexical style. To isolate the effect of data domain (not scale), we match the total training steps to our baseline CTS experiments, ensuring comparable optimization cycles. This setup tests whether cross-modal alignment generalizes to heterogeneous text distributions.

\begin{table*}[h!]
\centering
\footnotesize
\begin{tabular}{l|ccc|c|cccc}
\toprule
\textbf{Model} & \multicolumn{4}{c|}{\textbf{Pre-training}} & \multicolumn{4}{c}{\textbf{Fine-tuning (hrs)}} \\
& \multicolumn{3}{c|}{\textbf{Speech (hrs)}} & \textbf{Text} & \\
\cmidrule(lr){2-4} \cmidrule(lr){6-9}
& \textbf{ASR} & \textbf{ST} & \textbf{Total} & \textbf{(token | sentence)} & \textbf{3-Way} & \textbf{ASR-only} & \textbf{ST-only} & \textbf{Total} \\
\midrule
Whisper-large-v2 & $555$k & $126$k & $680$k & -- & -- & -- & -- & -- \\
NLLB-1.3B & -- & -- & -- & N/A | $> 40$B & -- & -- & -- & -- \\
SeamlessM4T-Large & N/A & N/A & $> 1$M & N/A | $> 40$B & N/A & N/A & N/A & $>400$k \\
STAC-ST & -- & -- & -- & -- & $206$ & -- & -- & $206$ \\
Bi-NMT & -- & -- & -- & -- & $206$ & -- & -- & $206$ \\
Multi-ST & -- & -- & -- & -- & $472$ & -- & -- & $472$ \\
Multi-ASR & -- & -- & -- & -- & $269$ & -- & -- & $269$ \\
QWen2-Audio & N/A & N/A & $> 5$M & $2.4$T | N/A & N/A & N/A & N/A & $520$k \\
\textbf{Whisper-UT (Ours)} & $555$k & $126$k & $680$k & -- & $110 \sim 180$ & -- & -- & $110 \sim 180$ \\
\bottomrule
\end{tabular}
\caption{Comparison of pre-training and fine-tuning data scales for Whisper-UT and baseline models. “3-Way Parallel” refers to datasets with aligned speech, transcripts, and translations. Note that "N/A" means some data is used for the specific training, yet the exact amount is not available.}
\label{tab:data_comparison}
\end{table*}

\subsubsection{Analysis of OOD Text Data Injection}
Injecting out‑of‑domain text under the unified objective appears to have limited benefit and in some cases even disrupted established behaviors. On Fisher, UT‑OOD (row 10) lags behind UT‑CTS across every translation metric—most notably MT accuracy, which jumps from 44.2 BLEU with OOD data to 55.9 BLEU when text is drawn from the CTS domain. This suggests that the linguistic and stylistic mismatch of web‑mined, TED talk, and parliamentary text fails to reinforce the speech‑to‑text alignment learned on conversational telephone speech, and may inject conflicting patterns that the model struggles to reconcile.

A similar story unfolds on BBN. On BBN, the impact of injecting OOD text is most pronounced in the MT task. Under UT‑OOD (row 21), the model’s MT performance barely improves over the base unified setting and remains far below the CTS‑matched variant—rising only to 11.1 BLEU compared with 15.7 BLEU for UT‑CTS (row 22). In contrast, UT‑CTS consistently lifts MT and MMT performance by several BLEU points and slightly improves ASR quality. Together, these findings imply that substituting in‑domain transcripts with heterogeneous text corpora does not generalize well in a cross‑modal training regime and can inadvertently weaken the model’s ability to leverage the unified translation objective.

\section{Model Training Data Overview}

Here, we also present a rough sketch of the training data amounts for our model and the compared methods, as summarized in Table \ref{tab:data_comparison}. However, it is important to note that due to differences in training methodologies, stages, and the unavailability of precise details for some systems, this comparison should be interpreted with caution and may contain ambiguities. We encourage readers to consult the original publications for more accurate and comprehensive descriptions of the training data used in each model.

\clearpage

\end{document}